\documentclass{article}

\PassOptionsToPackage{authoryear}{natbib}
\usepackage{acl}
\usepackage{times}
\usepackage{latexsym}

\usepackage[utf8]{inputenc} 
\usepackage[T1]{fontenc}    
\usepackage{hyperref}       
\usepackage{url}            
\usepackage{booktabs}       
\usepackage{amsfonts}       
\usepackage{nicefrac}       
\usepackage{microtype}      
\usepackage{xcolor}         
\usepackage[pdftex]{graphicx}

\usepackage{stfloats}

\RequirePackage{algorithm}
\RequirePackage{algpseudocode}

\usepackage{amsthm}
\newcommand{\RR}{\mathbb{R}}
\newcommand{\ZZ}{\mathbb{Z}}

\usepackage{multirow,tabularx}
\usepackage{footnote}
\makesavenoteenv{tabular}
\makesavenoteenv{table}
\makesavenoteenv{table*}
\usepackage{amsmath}
\usepackage{mathtools}
\usepackage{subcaption}
\usepackage{todonotes}
\usepackage{bigdelim}
\usepackage{capt-of}

\title{Sparsifying Transformer Models with Trainable Representation Pooling}

\author{%
  Michał Pietruszka\textsuperscript{1, 2} \quad Łukasz Borchmann\textsuperscript{1, 3} \quad Łukasz Garncarek\textsuperscript{1} \\
  \textsuperscript{1}Applica.ai \quad \textsuperscript{2}Jagiellonian University \quad \textsuperscript{3}Poznan University of Technology \\
  \texttt{\{michal.pietruszka, lukasz.borchmann,}\\ \texttt{lukasz.garncarek\}@applica.ai} \\
}

\begin{document}

\maketitle

\begin{abstract}
We propose a novel method to sparsify attention in the Transformer model by learning to select the most-informative token representations during the training process, thus focusing on the task-specific parts of an input. 
A reduction of quadratic time and memory complexity to sublinear was achieved due to a robust trainable top-$k$ operator.
Our experiments on a challenging long document summarization task show that even our simple baseline performs comparably to the current SOTA, and with trainable pooling we can retain its top quality, while being $1.8\times$ faster during training, $4.5\times$ faster during inference and up to $13\times$ more computationally efficient in the decoder.\footnote{Code publicly available at \url{https://github.com/applicaai/pyramidions} along with trained models.}
\end{abstract}

\section{Introduction}  

The introduction of Transformer architecture led to an immense improvement in the performance of Natural Language Processing systems \citep{Vaswani2017AttentionIA,Radford2018ImprovingLU,Devlin2019BERTPO}. Nevertheless, the underlying attention mechanism is marked by the original sin of quadratic memory complexity w.r.t. the input sequence length. It results from the attention matrix reflecting inter-connections between every two representations in the input sequence.

\begin{figure}
    \centering
        \includegraphics[width=\linewidth]{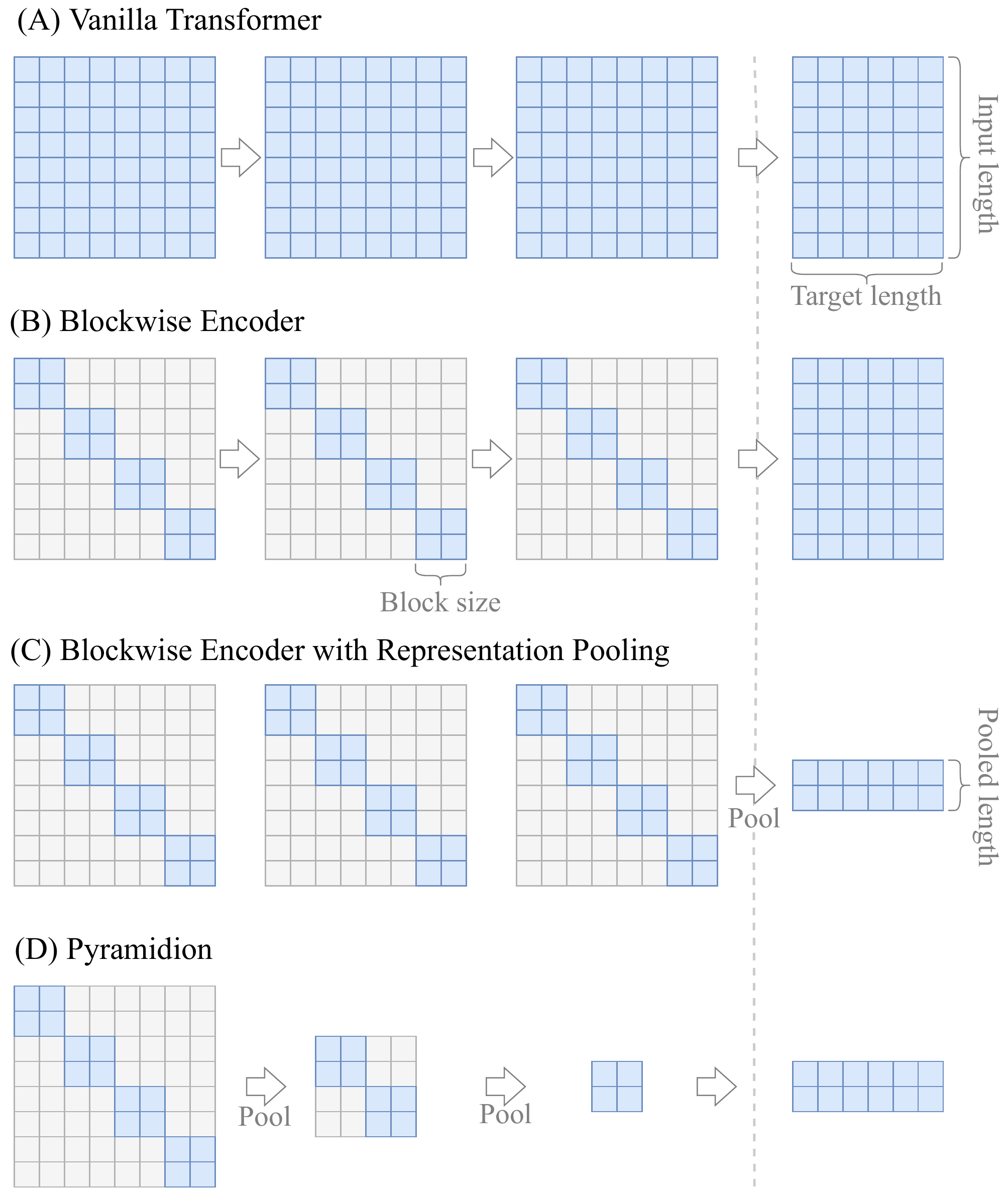}
        \caption{
        An illustration of sparse attention matrices assuming a three-layer encoder and decoder (separated by the dashed line). The blue color reflects the memory consumption of self-attention (encoder) and cross-attention (decoder).
        (A) The complete input consumed at once. 
        (B) Memory reduced with blockwise attention and (C) pooling applied after the encoder. (D) Gradual reduction of memory by pooling after every layer.}
        \label{fig:comparison}
\end{figure}

\begin{figure}
    \centering
        \vspace{-15mm}
        \centering
        \includegraphics[width=0.9\linewidth]{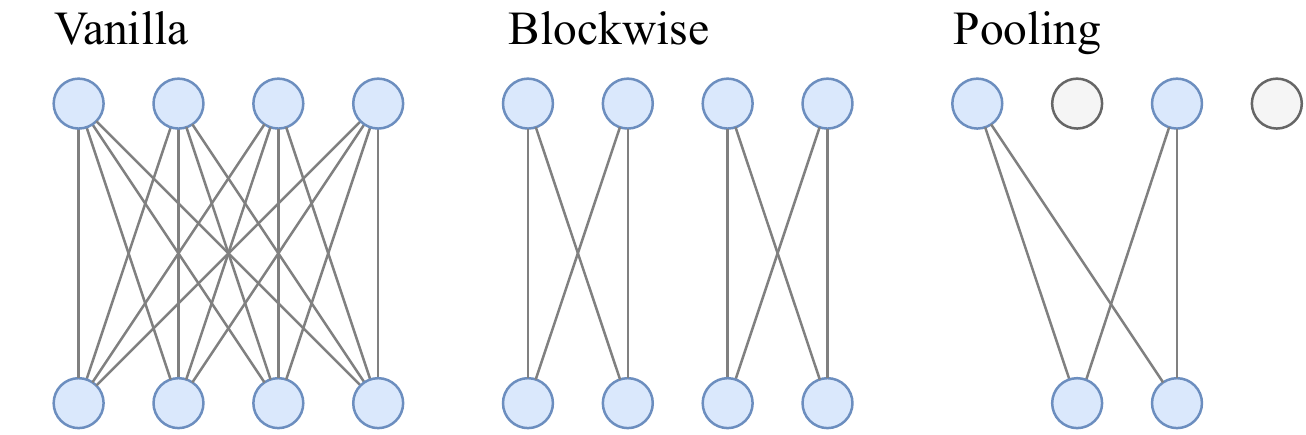}
        \caption{Toy illustration of inter-connections constituting the attention matrices in various approaches to attention. White dots denote disregarded representations that are not attended to and removed from further processing as they obtained low scores.}

        \label{fig:inter}
\end{figure}

\looseness=-1 Previous approaches either reduce the full connectivity of its elements to its non-empty subset or approximate the self-attention matrix \citep{Dai2019TransformerXLAL,Beltagy2020LongformerTL,Kitaev2020ReformerTE,tay2020sparse,zaheer2020big,wang2020linformer,shen2021efficient,choromanski2021rethinking,roy2020efficient}. In particular, in these models, each word at every layer attends to at least one other word.

In contrast, we disregard attention for a given representation completely in the case of non-informative ones~(Figure~\ref{fig:comparison} and \ref{fig:inter}).

In particular, we optimize the attention complexity by learning to select encoded representations for the given task and \textit{promoting} only the chosen ones to the next layer of the model. This mechanism will be referred to as \textit{representation pooling}. Consequently, a significantly lower memory consumption and an improved processing time are achieved. As the selection operation has to be trainable, we provide a suitable high-performance continuous relaxation of top-$k$, robust for every $k$ value and input sequence length. 

We demonstrate this idea's applicability by performing on par to state-of-the-art on the challenging problem of long document summarization. Simultaneously, the proposed end-to-end model is a significant theoretical improvement over the previous systems, which are based on independently trained extractive and abstractive models.

\paragraph{Contribution.} The specific contributions of this paper are the following: (1)~We propose a method to sparsify Transformer architecture in a novel, previously unrecognized way, achieving sublinear time and memory complexity. Our model learns to select the subset of best representations depending on the advantage they give on a downstream task.
(2)~Additionally, we demonstrate an improvement of the decoder's cross-attention complexity. It is beneficial for both train/inference time and memory consumption.
(3)~We demonstrate an elegant way to train extractive-abstractive models in an end-to-end manner with only a cross-entropy loss function. (4)~We present a Successive Halving Top-$k$ operator that outperforms previous approaches in terms of approximation quality and speed. We provide a detailed analysis of its differential properties and prove that it is trainable in an end-to-end manner, making it applicable within our neural networks. (5)~We achieve state-of-the-art performance level in long document's summarization and show that previous models can be outperformed by a straightforward baseline.

\section{Related Works}  
\paragraph{Word-vector elimination.}
It has been previously shown that the progressive elimination of word vectors occurring layer after layer can improve inference time of transformer-based language models used in a text classification scenario \cite{pmlr-v119-goyal20a}. We extend this notion to tasks demanding text generation in a way that, contrary to previous work, is trainable and optimized concerning a downstream task. A similar approach has been taken in the Funnel Transformer proposed concurrently to our work \citep{dai2020funneltransformer}. We directly compare to both methods' adaptations (see Section~\ref{sec:evaluation}), and consider our work to surpass it in two aspects: 1) results were improved due to a better pooling mechanism than mean/max; 2) training was accelerated, which we attribute to the significant reduction of the decoder's complexity.

\paragraph{Sparse attention.} 
Several authors proposed to limit attention connectivity, e.g., by dividing input into smaller 'blocks' \citep{child2019generating, Beltagy2020LongformerTL, rae-razavi-2020-transformers}. 
Blockwise attention is an optional element of our architectures, used in addition to trainable pooling.

\paragraph{Summarization.} 

In terms of the type of summarization task we target, our representation pooling mechanism can be considered an end-to-end extractive-abstractive model. This is a conceptual breakthrough compared to recently proposed two-stage hybrids that extract and paraphrase in two independent steps, using separately trained modules \citep{s2019extractive, hsu2018unified, gehrmann2018bottomup, chen2018fast}.

\section{Novel Approach of Representation Pooling}
It is suspected that when humans engage in information search, they use various cognitive processes depending on the relevance level of constituent text fragments \citep{asi.23904}.

The method we propose is inspired by this search for relevant fragments, which is an important aspect of human cognition when engaged in \textit{reading to do} actions 
\citep{mosenthal1996understanding,10.2307/40032191}. 
\looseness=-1 We intend to mimic relevance judgments and hypothesize that it is possible to answer problems involving natural language with only selected passages of the input text.

These passages may be of substantially shorter length than the original text.
One may compare this to a person reading the paper and highlighting in such a way that it is possible to provide a summary using only the highlighted parts.

\begin{figure*}
    \centering
    \includegraphics[width=0.9\linewidth]{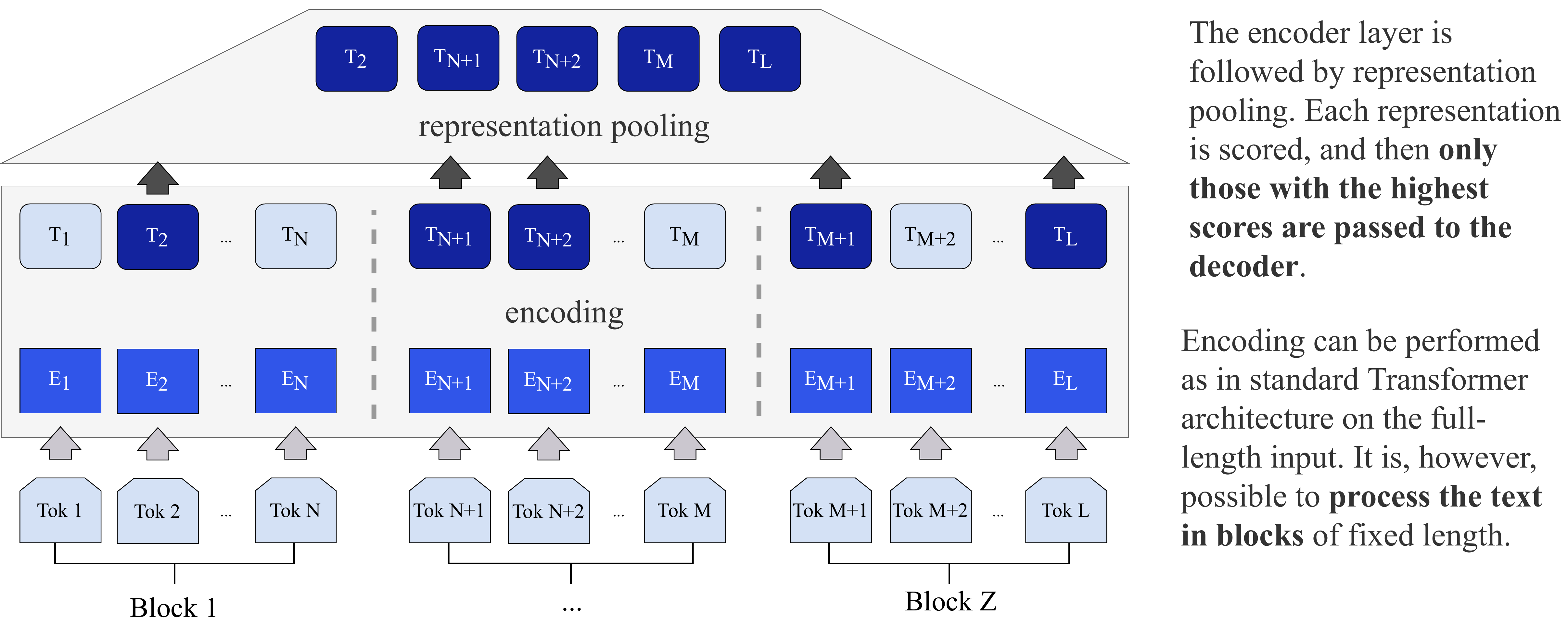} 
    \caption{
        Transpooler architecture with pooling after one encoder layer. Each representation is scored, and then only those with the highest scores are passed to the decoder. Encoding can be performed on the full length input or in blocks of fixed length.}
    \label{fig:representation}
\end{figure*}

\looseness=-1 The end-to-end mechanism we introduce performs such highlighting by 
scoring the representations and passes only the selected ones to the next layer of the neural network (Figure~\ref{fig:representation}). The role of the selection is to reduce data resolution in a roughly similar way to how pooling works in CNNs, where the feature map is downsampled and only the most informative activations are retained. When pooling in a trainable manner at the bottleneck of the encoder-decoder, it impacts the encoding process because the additional, orthogonal, informational bottleneck forces the model to compress more context into one representation vector of constant-length, leveraging the already provided capacity.

\subsection{Architecture Outline}
Let $n$ denote the number of input tokens that are projected onto $d$ dimensions, resulting in a matrix of embedding representations $\mathnormal{E \in \mathbb{R}^{n\times d}}$. We want to assign scores $v_{i}$ to embedding vectors $E_{i}$, in such a way that $v_{i}$ measures the usefulness of $E_{i}$ for further layers and the training objective.

Typically, this can be achieved by defining a scoring function $S\colon \mathbb{R}^{d} \to \mathbb{R}$ (which we allow to depend on additional parameters, thus making it trainable) that assigns a usefulness score to every embedding vector, and putting
\begin{equation}
  v_{i} = S(E_{i}).
\end{equation}

Next, we use our soft top-$k$ operator $\Gamma\colon \mathbb{R}^{n \times d} \times \mathbb{R}^{n} \to \mathbb{R}^{k \times d}$ to reduce the number of embeddings from $n$ to $k$, based on their usefulness scores. The $k$ vectors produced by $\Gamma$ form the input for the next network layer. The path of residual connections starts on a reduced number of tokens.

\paragraph{Flavors.} We consider two architectures in this work: with single or multiple pooling layers (Figure~\ref{fig:comparison}). Specifically, the latter is a generalization of the former to any given number of pooling layers. We use the term Transpooler when a single pooling layer is placed after the encoder. This setup directly limits the amount of information passed to the decoder through the network's bottleneck.

However, pooling can be applied between any subsequent layers, such that multiple operations of this type will be used in the network and gradually introduce the bottleneck along the encoding process. As a result, the same model bottleneck size can be achieved as when using Transpooler. Moreover, the decision to pool earlier has the advantage of attaining more substantial memory complexity reduction. This model will be referred to as the Pyramidion.

\paragraph{Blockwise attention.}
When propagating through layers, we use blockwise attention and split input into non-overlapping chunks in such a way that the full quadratic attention is computed for each chunk. The score is then determined for each representation vector, and after selecting with the top-$k$ operator, chosen representations are passed to the next layer. We assure our top-$k$ operator selects representations without permuting their order, keeping them in line with their original position. 

\paragraph{Scoring functions.\label{sec:scoring}} 
Multiple scoring methods can be proposed. The most straightforward is to use a linear scoring function as used in conventional token classification, $S(e)=e^T w+b$, where $w\in \mathbb{R}^{d}$ and $b\in\mathbb{R}$ are trainable parameters. We found it to work best with our pooling method. In the Appendix~\ref{ablations_scorers} we perform ablations on different scoring functions.

\subsection{Complexity Analysis}

\begin{table}
    \centering
    \caption{
        Time complexity of attention in the Transformer models. Improvements over the vanilla Transformer are in \textbf{bold}, whereas an \underline{underline} indicates this paper's contributions.
        $\mathnormal{l}$ -- number of layers, $\mathnormal{n}$ -- input length, $\mathnormal{d}$ -- hidden state;s size, $\mathnormal{t}$ -- target length, $\mathnormal{h}$ -- number of hashes LSH, $\mathnormal{r}$ -- rank of the factorization matrix, $\mathnormal{k}$ -- length of selected token's representation, $\mathnormal{c}$ -- an effective number of layers that is smaller than $\mathnormal{l}$.
        }
        \label{complexity}
        \centering
        \begin{tabular}{lcc}
        \toprule
        Model & Self-attention & Cross-attention \\ \midrule
        Vanilla  & l $\times$ n $\times$ n $\times$ d & l $\times$ t $\times$ n $\times$ d \\     
        Sparse& l $\times$ \textbf{m} $\times$ n $\times$ d & l $\times$ t $\times$ n $\times$ d \\
        Linformer  & l $\times$ n $\times$ \textbf{r} $\times$ d & --- \\
        LSH 
        & l $\times$ \textbf{mh} $\times$ n $\times$ d &  --- \\
        Efficient & l $\times$ n $\times$ \textbf{d} $\times$ d & --- \\
        PoWER & \textbf{c} $\times$ n $\times$ n $\times$ d  & --- \\
        \midrule
        Transpooler & l $\times$ \textbf{m}  $\times$ n $\times$ d 
            & l $\times$ t $\times$ \textbf{\underline{k}} $\times$ d \\
        Pyramidion & \textbf{\underline{c}} $\times$ \textbf{m} $\times$ n $\times$ d 
            & l $\times$ t $\times$ \textbf{\underline{k}} $\times$ d \\
         \bottomrule
         \end{tabular}
\end{table}

\looseness=-1  Table~\ref{complexity} presents the complexity of attention in our models, and compares it to different architectures. 
The vanilla encoder depends on the number of layers $\mathnormal{l}$, the number of tokens in the input $\mathnormal{n}$ and the number of tokens each attends to $\mathnormal{n}$. Likewise, the decoder's cross-attention depends on $\mathnormal{l}$, $\mathnormal{n}$ and the target length $\mathnormal{t}$.

\looseness=-1 The $m$ denotes the effective number of tokens one can attend to, resulting from the attention's block size, allowed window size or the clustering of key-values. The number of parallel LSH hashes is denoted by $\mathnormal{h}$. The rank of the factorization matrix is $\mathnormal{r}$, which can be a constant that is independent of $\mathnormal{n}$.

Similarly, the number of best task-specific representations $\mathnormal{k}$, selected after encoding, is independent of $\mathnormal{n}$. $\mathnormal{c}$ is an effective number of layers in a hierarchically decreasing encoder of the Pyramidion. The Pyramidion's $\mathnormal{c}$ can be as low as $\mathnormal{2}$.
Blockwise sparse attention improved the vanilla Transformer's complexity by limiting the number of tokens each attends to from $\mathnormal{n}$ (input length) to $\mathnormal{m}$ (block size) as seen in Table~\ref{complexity}. As we keep the encoding of blockwise attention, the $m$ improvement also applies to our self-attention. 

For the Pyramidion model, we narrow down the size of the representation on the output of each chosen layer, leading to the exponential reduction of memory consumption as the encoding proceeds. For example, when pooling after every layer is considered, the total memory complexity across $l$ layers would be $
    \sum_{i=0}^{p} 2^{-i}  m n d
    =   (2-k/n) m n d
$
where $p$ denotes the number of passes $p = \log_2(n/k)$, assuming $k\leq n$ and $n, k \in \{2^i \mid i \in \ZZ_{+}\}$.
Hence, the effective complexity of all layers is lower than $\mathnormal{2 m n d}$, which means it is lower than $\mathnormal{2}$ times the complexity of the full-size first layer.

For the decoder cross-attention, the number of input representations that $t$ target tokens can attend to is limited by $k$, thus decreasing the memory complexity of cross attention from  $\mathnormal{\mathcal{O}(t n)}$ to $\mathnormal{\mathcal{O}(t k)}$. Optimization over quadratic sentence-length complexity is even more powerful and needed on the decoder side, as $\mathnormal{\mathcal{O}(t n)}$ complexity hurts performance of real-world applications based on auto-regressive decoding.

The blockwise attention itself reduces encoder complexity proportionally to the number of chunks.
We further reduce the decoder layer's complexity in Transpooler models by a factor of $n/k$, thanks to representation pooling. The Pyramidion we propose offers an additional improvement on the encoder side, where time and memory consumption are reduced in each of the consecutive layers compared to the Transformer featuring blockwise attention. In other words, when $b$ denotes the number of blocks, $l$ stands for the number of layers, and the sequence length is halved in each layer, we reduce memory from $b + b + ... + b = lb$ to $b + b/2 + b/4 + ... + b/(2^l) \leq 2b$.
Because the beneficial impact of pooling accumulates, we are able to improve complexity from one that is linearly dependent on $l$ to one that is constant, independent of $l$. In the further DeepPyramidion's experiments, we will proceed with a higher reduction factor, where the length of a sequence is cut in four.

\begin{figure}
        \centering
        \includegraphics[width=\linewidth]{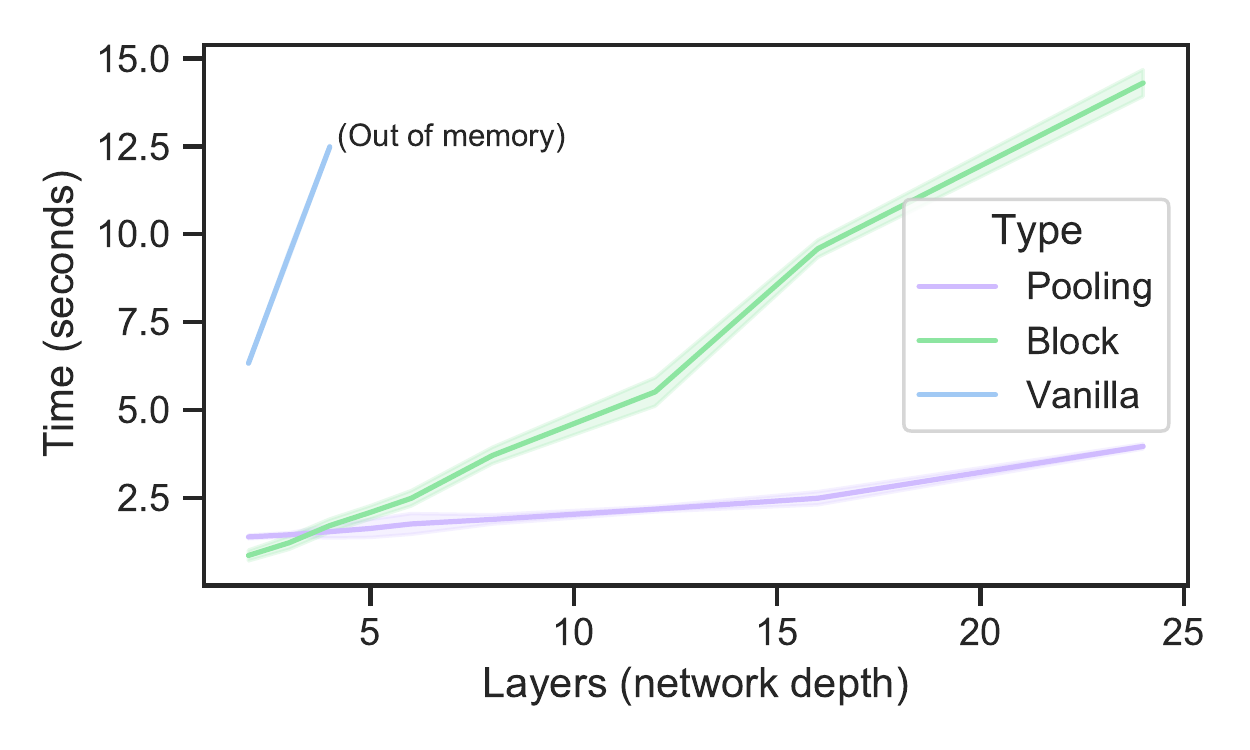}
        \caption{Training time for different model sizes of Vanilla Transformer, Blockwise, and Pyramidion $8k\rightarrow512$ with the input sequence length of $8192$ tokens.
        Pooling is faster for models with $4$ or more layers, achieving up to $3.8x$ speedup for $16$-layer Transformer. Scores of a $2$-layer version of these models do not differ significantly.
        }
        \label{fig:performance_benchmark}
\end{figure}

As a result, the Pyramidion achieves an effective self-attention time and space complexity linear of $n$ and logarithmic of $l$. 
For comparison, other sparse models such as, e.g., Linformer depend linearly on $n$ and linearly on $l$. The analysis of Figure~\ref{fig:performance_benchmark} found evidence that our method scales well with an increasing number of layers. In the evaluation (see Section \ref{sec:evaluation}), we demonstrate that our model achieves a $2.5\times$ computation reduction in the encoder's self-attention and a $16\times$ reduction in the decoder's cross-attention comparing to blockwise baseline, while both models are close to SOTA results on the task of long-document summarization.
All things considered, we introduce Pyramidion with sublinear complexity that achieves remarkable results.

The advantage of our approach is that it complements all other proposed sparsification techniques, thus paving a new interesting avenue of potential research. 
It can be effortlessly applied in-between layers and simultaneously with other improvements since representation pooling addresses a different aspect of the attention's complexity problem.

\section{Suitable Top-\textit{k} Operator} \label{sec:topk}
The choice of the selection operator is challenging, as it has to be trainable to instantiate a pooler.
In case of the hard top-\(k\) operator,  back-propagation through the scores is impossible and prevents training the scoring function.  It could be seen as an extreme case of the vanishing gradient problem. In this section we introduce a mechanism not prone to this issue, while the Appendix~\ref{appendix_topk} is dedicated to a theoretical analysis of its differential properties, from a geometrical point of view.

The crux of our approach is the Successive Halving Top-$k$ selection mechanism
that finds $k$ convex combinations of vector representations $E_{i}$, dominated by those achieving the highest scores $v_{i}$ (pseudocode available in the Appendix \ref{sec:limitations}).\footnote{Preliminary work regarding this method was previously presented in the form of a Student Abstract, see \citet{pietruszka2020successive}.} The general idea is to perform a tournament soft selection, where candidate vectors are compared in pairs $(i, j)$, until only $k$ remained. After each tournament's round new $E'$ and $v'$ are computed as convex combinations of these pairs with weights based on their respective scores. Each new vector is calculated as:
\begin{equation*}
    E'_{i} = w_{i}E_{i} + w_{j}E_{j},
\end{equation*}
where the $w_i, w_j$ are the result of a peaked softmax over the scores $v_i, v_j$.
Analogously, we use $v'_{i} = w_{i}v_{i} + w_{j}v_{j}$ as the new-round's scores.

Weights are calculated using a $\operatorname{PeakedSoftmax}$ function \citep{goyal-etal-2017-differentiable}, increasing the pairwise difference in scores between $v_i$ and $v_j$. 
One round halves the number of elements in $E$ and $v$. We perform it iteratively unless the size of $E$ and $v$ matches the chosen value of $k$. 

To improve convergence towards selecting the real top-$k$, it is desired to permute $v$ and $E$ first. In our algorithm, we sort the vectors $E_{i}$ in descending order of their scores $v_{i}$ and then put them into the tournament in pairs of the form $(i, n+1-i)$. This method of pairing guarantees that the weights $w_i$ depend monotonically on the scores $v_i$, which is the main motivation for using it. Extended benchmarks for time and accuracy are covered in details in  Appendix~\ref{topk_performance}.

\begin{table*}[ht!]
\caption{Scores, complexity and benchmark depending on maximum encoder and decoder lengths, as well as used sparsification mechanism. All models features a $2$-layer encoder and a $2$-layer decoder, blocks of size $512$. Results on arXiv summarization dataset \citep{cohan-etal-2018-discourse}. Arrow $\rightarrow$ denotes a pooling operation additional to the one between encoder and decoder. Note, that for the vanilla Transformer encoder lengths are equal to the decoder's length, whereas Transpoolers and Pyramidions lower the number of representations passed down to the decoder without the substantial quality decrese.}
    \label{tab:length}
\centering
    \begin{tabular}{rrrrrrrr}
    \toprule
    \multirow{2}{*}{\#} &
    \multirow{2}{*}{Architecture} & 
    \multicolumn{2}{c}{Lengths} &
    \multicolumn{2}{c}{Time} &
    \multicolumn{2}{c}{ROUGE} \\
    & & Encoder & Decoder  & Training & Inference & R-1 & R-2 \\
    \midrule
    1 & \ldelim\{{3}{30pt}[Vanilla ] & 512 & 512  & 0.13 & 4.23 & 28.1 & 8.3 \\
    2 & & 2k  & 2k & 0.60 & 5.77 & 38.2 & 14.0 \\ \vspace{0.1cm}
    3 & & 8k  & 8k & 4.46 & 13.27 & \textbf{41.8} & \textbf{16.1} \\ 
    4 & \ldelim\{{2}{44pt}[Blockwise ] & 2k & 2k & 0.31 & 5.28 & 38.6 & 14.1 \\ \vspace{0.1cm}
    5 & & 8k & 8k & 0.85 & 11.49 & \textbf{41.9} & \textbf{16.7} \\ 
    6 & \ldelim\{{3}{50pt}[Transpooler ]  & 2k  & 512 & 0.54 & 4.24 & 39.1 & 14.6 \\
    7 & & 8k & 512 & 1.44 & 4.28 & 41.8 & 16.4 \\ \vspace{0.1cm}
    8 & & 8k  & 2k & 1.26 & 5.51 & \textbf{42.7}& \textbf{16.7}\\
    
    \midrule
    9 & \ldelim\{{3}{105pt}[LSH \citep{Kitaev2020ReformerTE} ] & 512 & 512 & 0.19 & 4.27 & 28.5 & 7.5 \\
    10 & & 2k &  2k & 0.56 & 5.92 & 33.6 & 10.5 \\ \vspace{0.1cm}
    11 & & 8k &  8k & 1.69 & 13.41 & 35.7 & 11.2 \\
    12 & \ldelim\{{3}{110pt}[Efficient \citep{shen2021efficient} ] & 512 & 512 & 0.12 & 4.20 & 28.4 & 7.8 \\
    13 & & 2k & 2k & 0.29 & 5.91 & 34.1 & 10.4 \\ \vspace{0.1cm}
    14 & & 8k & 8k & 0.82 & 13.75 & 35.0 & 10.8 \\
    15 & \ldelim\{{3}{115pt}[PoWER \citep{pmlr-v119-goyal20a} ] & 2k $\rightarrow$ 1k & 512 & 1.04 & 4.28 & 35.3 & 12.7 \\
    16 & & 8k $\rightarrow$ 2k & 512 & 1.87  & 5.33 & 36.9 & 14.1 \\ \vspace{0.1cm}
    17 & & 8k $\rightarrow$ 4k & 2k & 2.06 & 6.92 & \textbf{42.0} & \textbf{16.5} \\
    18 & \ldelim\{{3}{100pt}[Funnel \citep{dai2020funneltransformer} ] & 2k $\rightarrow$ 512 & 2k & 0.61 & 4.01 & 38.6 & 14.3 \\
    19 & & 8k $\rightarrow$ 512 & 8k & 1.78  & 4.03 & 41.8 & \textbf{16.5} \\ \vspace{0.1cm}
    20 & & 8k $\rightarrow$ 2k & 8k & 1.53 & 5.25 & \textbf{42.0} & 16.4 \\
    \bottomrule
    \end{tabular}
\end{table*}

\section{Evaluation} \label{sec:evaluation}
The main focus of the experiments was to understand how to employ the Successive Halving Top-$k$ operator within neural networks to build models that have better training and inference time and are expressive enough to achieve results comparable to state-of-the-art models. The first experiment was specifically designed to compare to other sparse Transformers and Vanilla baselines.

\paragraph{Choice of tasks.} We demonstrate the benefit of pooling on the arXiv and PubMed summarization datasets \cite{cohan-etal-2018-discourse} available under Apache License 2.0 license. Both tasks demand text generation and have the highest average input sequence length ($6$k and $3$k words on average for arXiv and PubMed respectively). Assuming an embedding of dimensionality $768$, it is important to note that for inputs shorter than approx. $4k$ tokens, more multiplications happen in the Transformer's FFN layers and projection layers than in the attention layers. Hence, the validation of the sparsification mechanism should be proved by showing that it works for longer inputs.

\paragraph{Time benchmarks.} The average time of processing a batch of documents is reported to evaluate the computational improvements experimentally. Decoding experiments were synthetic with a forced fixed length of $512$ output tokens to discount for the lower processing time of models predicting an earlier sequence end.
We recorded time in seconds on batches of size $64$ and $8$ for training and generation, respectively. Details regarding the hyperparameters and test environment are reported in Appendix~\ref{appendix_experiment_details}.

\paragraph{Ablations on input and decoder lengths.} Table~\ref{tab:length} presents evaluation metrics and time benchmarks depending on encoder and decoder lengths, as well as used sparsification mechanisms. At this stage, we use shallow $4$-layer models to perform ablation studies and estimate each approach's strengths and weaknesses. We observe that all sparse models deliver on the promise of accelerating training time over Vanilla Transformers for longer sequences in this setup. Methods requiring the elimination of word vectors scale well with the sequence length but incur additional pooling costs, which may be notable for shorter sequences.
Nevertheless, inference time was significantly reduced only when methods eliminating word vectors were employed. The introduction of blockwise attention and pooling does not decrease scores while lowering the computational cost. The detailed training procedure for all models is provided in Appendix~\ref{appendix_experiment_details}.

\paragraph{Scaling deeper.} In preliminary experiments it was estimated that the fastest-to-train model that performs comparably to the Vanilla Transformer is the Blockwise Transformer. Here, we scale it to $6$-layers in each encoder and decoder and provide an interesting baseline for our model, since Transpooler's backbone is blockwise attention. 
We undertook the empirical analysis of scaling Transpooler to many layers in Appendix~\ref{appendix_sub_pyramidion} and found that in order to balance performance and speed, it is crucial to delay the first pooling and not to perform it directly on the first layer's output. It was also revealed that appending more layers at the end of the encoder (after pooling) results in a negligible increase in time while considerably improving scores. Both changes to the block size and reduction of the bottleneck harmed the performance. Thus, the data supports the premise that the $6$-layers encoder should consume $8k$ tokens on the input and output representations of lengths $8k,8k,2k,512,512,512$ after each successive layer. We refer to this model as DeepPyramidion (note that pooling happens twice in the encoder). The decoder also has six layers, making our model directly comparable to the deeper Blockwise Transformer. We confront DeepPyramidion with the Blockwise baseline by training models from scratch on arXiv and PubMed datasets separately and report results in comparison to the state-of-the-art summarization models (Table~\ref{tab:sota}).

\paragraph{Results.} 
\looseness=-1 The evaluation of the data presented in Table~\ref{tab:sota} leads to the unexpected conclusion that our Blockwise Transformer baseline, despite its simplicity, is sufficient to outperform deeper, denser, and additionally pretrained models that were recently reported as state-of-the-art. We demonstrate that DeepPyramidion retains or improves the performance of the competitive baseline we produced. The training time speedup by $1.8\times$ supports the notion that our model scales better to long sequences, assuming deeper models. This result stands in line with evidence in Figure~\ref{fig:performance_benchmark}. While our baseline Blockwise model reduces the computational demand of self-attention in encoder by a factor of $16\times$ when comparing to Vanilla Transformer, it does not improve the decoder's computational complexity. It is interesting to highlight that DeepPyramidion further lowers the cost of self-attention by $2.5\times$ and improves $16\times$ over Blockwise's cross-attention in the decoder, and leads to overall $13\times$ improvement in the number of multiplication operations in the decoder.  Time benchmarks show a $4.5\times$ improvement in the generation times for our method, proving how vital the improvement in the decoder's cross-attention complexity is for inference time.

\looseness=-1 DeepPyramidion achieves a ROUGE-2 score indistinguishable from SOTA on arXiv and performs competitively on PubMeb. At the same time, an entire DeepPyramidion costs five times less than a single Transformer layer consuming $8k$ tokens. However, when comparing our results to those of older studies, it must be pointed out that our models were trained from scratch only on the targeted dataset, whereas prior works often base on already pretrained models such as BART or RoBERTa and leverage unsupervised training on additional datasets. On the contrary, a longer input sequence was consumed by both Blockwise and DeepPyramidion, which we speculate, is the reason for their strong performance.\footnote{\looseness=-1 This view is supported by results of PoolingFormer that are concurrent to our work \citep{zhang2021poolingformer}. Despite that, at first sight, the methods seem similar and the authors present an interesting use of pooling in the attention, we argue that the mentioned model suffers from several weaknesses that are not present in our work. First of all, in the PoolingFormer model vectors are not removed from computations in further layers. Hence logarithmic complexity of the number of layers does not apply. PoolingFormer's approach suffers from having three orders of magnitude more calculations than when a global pooling based on scores of individual tokens is considered.}

\paragraph{Impact of longer inputs.}
The results achieved in our paper are comparable to other, much heavier, and more costly models due to two main reasons, that will be briefly discussed below.

Firstly, to perform well on a long document summarization task, there is a need to strike the right balance not only between the depth and width of the network but also it is required for design optimization to take into account the length of the input. All previous work seem to underperform when considering all three factors, as they were designed and optimized for shorter tasks and generally have more parameters, denser computations, or even a hard limit on the range of positional encoding. The authors were thus bounded by the maximal sequence length of $512$ or $1024$ tokens. One can argue that within this prefix (corresponding to the first $2-3$ pages), any data point from the arXiv/PubMed datasets (a scientific paper) usually provides enough information to write a meaningful summary, but also, important details will be missing to some degree. Hence, increasing the length of the input that can be consumed on GPUs, at the price of using a shallower network, with sparser computation, may be considered a better fit for the task.

Secondly, we think that pretraining in the Pyramidion’s case may be disregarded due to an interesting “length exploiting hypothesis”. That is, while we consume longer sequences on the input, the network learns more efficiently, as more information is available, and thus, the training signal is stronger. This can be convincingly portrayed in the case of embedding layers, as during training they see many more words and sentences from the chosen dataset, and hence, can provide more meaningful representations to the further layers.

One can think that making the most of already available domain texts and consuming longer inputs is an advantageous approach to masked pretraining on out-of-domain datasets. While the latter approach may aid ‘general’ language understanding, it has insufficient transferability potential to domain-specific document understanding (e.g.,  scientific or medical texts).

To sum up, the Pyramidion has improvements that allow consuming longer inputs cheaply, which turns out to be a more cost-effective strategy compared to other models. This aspect is crucial for achieving strong results on the presented datasets.

\begin{table*}[t]
    \caption{Comparison to SOTA on long document summarization tasks.
    Our models have no pretraining whereas $\dagger$ were initialized from BART, $\ddag$ -- from RoBERTa, $^*$ -- from PEGASUS  \citep{zhang2021poolingformer,rohde2021hierarchical,zaheer2020bigbird,gidiotis2020divideandconquer}.}
    \label{tab:sota}
    \centering
    \begin{tabular}{lrrrrrrrr}
    \toprule
    \multirow{2}{*}{Architecture} &
    \multicolumn{2}{c}{arXiv} &
    \multicolumn{2}{c}{PubMed} &
    \multirow{2}{*}{Params} &
    \multicolumn{2}{c}{Time}
    \\
    & R-1 & R-2 & R-1 & R-2 & & Train. & Infer. \\
    \midrule
    PoolingFormer$\dagger$ & \textbf{48.47}& \textbf{20.23} & -- & -- & $>$406M  & -- & -- \\
    HAT-BART$\dagger$ & 46.74 & 19.19 & \textbf{48.25} & \textbf{21.35 }& $>$406M & -- & -- \\
    BigBird-PEGASUS$\ddag$ & 46.63 & 19.02 & 46.32 & 20.65 &  568M & -- & -- \\
    Dancer PEGASUS$^*$ & 45.01 & 17.60 & 46.34 & 19.97 & 568M & -- & -- \\
    \midrule
    Blockwise (our baseline) & 46.85 & 19.39 & -- & -- & 124M  & 4.85 & 37.15\\
    DeepPyramidion (our) & 47.15 & \textbf{19.99} & 47.81 & \textbf{21.14} & 124M & 2.71 & 8.12 \\
    \bottomrule
    \end{tabular}
\end{table*}

\section{Limitations and Social Impact}\label{sec:limitions_and_social_impact}

At this stage of understanding, we believe that sparsification based on trainable pooling is unlikely to improve processing time for short sequences specific to some NLP tasks, e.g., sentence-level Neural Machine Translation. In addition, the score improvement may be attainable for tasks characterized by at least an order of magnitude shorter outputs than inputs, as it was previously shown on classification, or, as in the case of this work, on summarization.

However, the extent to which it is possible to replace full-attention in Transformer with the sparse attention we propose is unknown. However, we argue that the benefits are visible starting from the inputs of length $4k$. As discussed earlier, $4k$ is a break-even point where more calculations are needed for attention than for FFNs and projecting layers. As such, we recommend applying sparsification methods on datasets featuring sequences of length over that value.
While we focus on the long end of the possible inputs, one can continue our analysis, to find improvements that work for shortest sequences, such as, e.g., concentrating on employing lighter projection layers and FFNs or stacking more attention blocks.

Although our method is a hybrid extractive-abstractive, it does not provide interpretable explanations to which specific representations were selected as the pooling operates in the latent space. How to match the selected vectors to the vocabulary tokens remains an open question. 
Moreover, framing the trainable pooling for language modeling remains a challenge to address in future works, especially as in this task the Markov assumption may serve as a basis for competitive pooling heuristics.

We did not consider Relative Positional Encoding in our work as pooling mechanism
is not trivially applicable with it and some generalization of our method may be needed. In that case, as it demands more experiments and proofs, we will leave the generalization of the pooling method for future work.

Regarding the social impact and environmental sustainability, we actively considered the Earth's well-being by contributing a technique for reducing the computational demand of recent Deep Learning models.
Our near-state-of-the-art DeepPyramidion model costs us $3$ days of training on $8$ NVIDIA A100 GPUs. Shallow models featuring trainable pooling were finished in about $2$ days each, given the same hardware. Blockwise baselines cost us about $3.5x$ the price of respective pooling methods. The most prolonged training of the $8k$ Vanilla Transformer lasted for about $2$ weeks. The total cost of training the models covered in this paper is about $2$ months on the mentioned hardware, plus an additional month for models and ablations described in the appendices.

We roughly estimate that it is between half and one-fourth of the total computation spent, including false runs, unpublished work, and initial experiments. The dataset preparation took less than $10$ hours on $1$ CPU.

\section{Summary}
We propose representation pooling as a method to reduce the complexity of Transformer encoder-decoder models. 
Specifically, we optimize self-attention complexity and address the decoder's cross-attention complexity optimization, which has so far not been widely acknowledged by the research community.
Moreover, the DeepPyramidion we introduced establishes results comparable to state-of-the-art, outperforming not only other systems relying on progressive word-vector elimination but also deeper, denser, and additionally pretrained models.

We tackle the problem by introducing a novel method of applying successive halving to a model's input in a tournament style. It is a theoretical improvement over existing approaches in terms of both computational complexity and approximation quality. 
Trainable Top-k selection allows to train scorer for a task and outperforms other pooling methods.

From the summarization task's point of view, the proposed end-to-end model is a significant theoretical improvement over the previous systems, where the extractive model was trained independently of the abstractive one. In contrast, our mechanism does not require the introduction of an additional training objective or training stage. 

Our approach can be easily applied to other problems from Natural Language Processing and Computer Vision. E.g., in a recent work later than ours, Multiscale Vision Transformers were proposed. These, similarly to our Pyramidion model, introduce the bottleneck gradually along the encoding process of videos and images, leading to better results, and complexity \cite{DBLP:journals/corr/abs-2104-11227}. As it comes to Natural Language Processing, possible applications include Key Information Extraction, Machine Reading Comprehension, and Question Answering in scenarios where encoder-decoder models struggle or would struggle with input sequence length (see, e.g., \citet{choi-etal-2017-coarse,townsend2021doc2dict,DBLP:journals/corr/abs-1712-07040}). We are looking forward to seeing these opportunities exploited.


\section*{Acknowledgments}

For easy reproduction of the results, we release our \linebreak utilities, code and pretrained models on the MIT li-\linebreak cense for all researchers not affiliated or wor-\linebreak king for Russian state-controlled institutions and \linebreak public companies. The reason to ostracize scientists \linebreak under those affiliations is the violent invasion of \linebreak their armed forces on Ukraine, recklessly intended to\linebreak inflict pain, threaten world peace and civilians life with nonhuman aggression against a sovereign nation.

The authors would like to thank Zofia Prochoroff and Paweł Morawiecki for the helpful discussions on the draft of the paper. Moreover, we thank the reviewers for their comments and suggestions that helped improve the paper.

The Smart Growth Operational Programme supported this research under project no. POIR.01.01.01-00-0877\linebreak/19-00 (\textit{A universal platform for robotic automation of processes requiring text comprehension, with a unique level of implementation and service automation}).
\bibliography{pyramidions}

\appendix
\clearpage
\newpage


\section{Scorers' Ablations}\label{ablations_scorers}

\paragraph{Linear.} 
Multiple scoring methods can be proposed. The most straightforward is to use a linear scoring function used in conventional token classification, $S(e)=e^T w+b$, where $w\in \mathbb{R}^{d}$ and $b\in\mathbb{R}$ are trainable parameters.

\paragraph{Nonlinear.} 
A quite natural next step is to include nonlinearity. We follow the specification of RoBERTa's classification head \citep{liu2019roberta}, defined as $S(e)=\tanh (e^T w_1+b_1)\cdot w_2 + b_2$, where $w_{1}, w_{2}\in \mathbb{R}^{d}$ and $b_{1}, b_{2}\in\mathbb{R}$.

\paragraph{PoWER-like.} 
A column-wise sum over attention matrices $A=\operatorname{Attn}(E)$ from the preceding layer can be used as the usefulness score, that is $v_{i} = \sum_{j=1}^{n} A_{i,j}$ as proposed by \citet{pmlr-v119-goyal20a} for hard top-$k$ selection.
\paragraph{Embedding-based.} 
Scoring can be performed based on a specified dimension in encoded space, i.e.\ by using a coordinate projection $S(e)=e_{j}$, where $j$ is a fixed index. This is a special case of the linear scoring function with fixed non-trainable weights.
\paragraph{Random.} The baseline sampling scores randomly from a uniform distribution.
\paragraph{Index-based.} A modulo-distributed score, that is non-zero for every $k$-th token, such as:
\[
v_i=\left\{
\begin{array}{l l}
1 & \mathrm{when}\ i \equiv 0 \pmod k \\
0 & \mathrm{otherwise} \\
\end{array}\right.
\]
\paragraph{Mean/Max Pooling.}
Pooling baselines characterized by aggregating scores within each window either by taking the mean value or the max value. In this case 4 nearest tokens were aggregated, and the window also traverse with the stride of 4.

Both the PoWER-like and embedding-based scoring functions utilize mechanisms already provided in the Transformer model and are easy to use. Similarly to the index-based baseline method and the random one, they do not introduce any additional parameters to the model. The last two do not rely on a pooling operation at all.

PoWER was proposed assuming that the model's attention already contains useful information about the most critical parts of the input sequence~\citep{goyal2017continuous}. In principle, it is  possible to use its scorer with soft top-$k$, but we intended to follow the original formulation where scoring was followed by the hard top-$k$ operation.

\subsection{Results} Results obtained with the same, $4$-layer Transpooler but different scoring functions are presented in Table~\ref{tab:scorer}.

All of the methods outperform the random baseline. Across them, the linear scorer achieved the highest evaluation metric. The index-based method we propose performs well, even though it does not require training.

In particular, models employing such fixed selection achieve better results than those equipped with a PoWER-like scorer. This can be attributed to the relatively low reduction of length required in the presented experiment: a model with index-based selection presumably learned to compress groups of the four nearest token neighbors.

Nevertheless, only nonlinear baseline approaches turned out not to be significantly worse than the linear scorer. Assuming preference towards a simpler method, the rest of the experiments were conducted using only the linear scorer.

\begin{table}[h]
\caption{Ablation study of different scorers, using the same $4$-layer Transpooler model with reduction from $2048$ to $512$ representations. 
The difference of $0.4$ is significant. \citep{doi:10.1113/jphysiol.2012.239376}.}

\label{tab:scorer}
\centering
\begin{tabular}{lcc}
\toprule
Scorer & ROUGE-1 & ROUGE-2 \\ \midrule
%
Linear     &  $\textbf{39.1}$ & $\textbf{14.6}$ \\
Nonlinear       & $\textbf{38.9}$ & $\textbf{14.6}$ \\        
%
%
Random         &  $32.3$ & $11.4$ \\        
Index-based          & $38.2$ & $13.9$ \\
Embedding-based &  $37.6$ & $14.0$ \\
PoWER-like &  $36.9$ & $13.6$  \\

Mean Pooling &  $38.1$ & $13.9$  \\

Max Pooling &  $38.4$ & $14.2$  \\

\bottomrule
\end{tabular}
\end{table}

\section{Successive Halving Top-\textit{k} Algorithm}\label{appendix_topk}
\citet{goyal2017continuous} provides the most similar relaxation for beam search, where they continuously relaxed the top-$k$-argmax procedure by performing softmaxes iteratively $\mathnormal{k}$ times and masking the previously extracted values. Each beam can contribute to the newly selected beam in every iteration, based on its distance to the max value.
By replacing one-hot coded vectors with their expectations in a similar vein, \citet{pltz2018neural} relaxed the KNN hard top-$k$ selection rule. 
\citet{xie2019reparameterizable} replaced a sampling of $k$ elements from the collection of items with Gumbel trick. 
Nevertheless, all the mentioned top-$k$ approaches remain too costly as they perform many iterations over a considered vector. Their time performance degrades due to $k$ softmaxes over the entire input length of $n$.

\citet{xie2020differentiable} parametrized the top-$k$ operator in terms of an optimal transport problem.  Employing such an algorithm instead of softmax may induce numerous zero weights in the attention matrix. However, this does not reduce the computational complexity of attention, as full-matrix multiplication has to be performed anyway and we are not concerned with such a method.

\subsection{Limitations and Assumptions\label{sec:limitations}}

The choice of the selection operator is challenging, as it has to be trainable to instantiate a pooler. Let us view the hard top-$k$ operator from a more geometric perspective.

In our setting, we consider sequences of \(n\) vectors from some vector space \(X\) (token embeddings), accompanied by real-valued scores, which are the basis for choosing the best \(k\) among \(n\) vectors. Thus, formally, a top-\(k\) operator should be defined as \(\Gamma\colon X^{n}\times \RR^{n}\to X^{k}\), assigning to a sequence of \(n\) vectors \(x_{i}\in X\) and their scores \(v_{i} \in \RR\) a sequence of \(k\) vectors \(y_{i}\in X\). For \(\Gamma\) to deserve the name `top-\(k\) operator', the output vectors \(y_{i}\) should depend \emph{mostly} on the \(k\) input vectors \(x_{i}\) with the largest corresponding scores.

In case of the hard top-\(k\) operator \(T\), the \(y_{i}\) are simply the vectors \(x_{i}\) with the largest scores, i.e.
\begin{equation}
  T((x_{i}), (v_{i})) = (x_{i_{1}}, x_{i_{2}}, \dots, x_{i_{k}}),
\end{equation}
where the indices \(i_{*}\) are chosen so that \(v_{i_{1}}\geq v_{i_{2}} \geq \dots \geq v_{i_{k}} \geq v_{j}\) for all \(j\not\in \{i_{1}, \dots, i_{k}\}\). In other words, \(T\) can be described as a composition of sorting the sequence \((x_{i})\) according to descending scores \(v_{i}\), and projecting onto \(X^{k}\) by discarding all but the first \(k\) elements.

To discuss the properties of \(T\), let us denote by \(S_{n}\) the set of all permutations of \(n\) indices \(\{1, 2, \ldots, n\}\). For every sequence \((x_{1}, x_{2}, \dots, x_{n})\) of length \(n\) there exists a permutation \(\sigma \in S_{n}\), such that $(x_{\sigma(1)}, x_{\sigma(2)}, \dots, x_{\sigma(n)})$ is sorted in descending order. We will refer to $\sigma$ as the \emph{sorting permutation} of the sequence $(x_i)$. It is unique, provided that the elements \(x_{i}\) are all distinct. Otherwise, the sequence \(x\) is invariant under permuting the indices of elements which are equal, and every two sorting permutations differ by such a factor.

For a permutation \(\sigma\in S_{n}\), define \(R_{\sigma}\subset \RR^{n}\) as the set of all vectors \(v\in \RR^{n}\) for which \(\sigma\) is a sorting permutation. The regions \(R_{\sigma}\) cover \(\RR^{n}\) and have disjoint interiors, containing vectors with pairwise distinct coordinates. The restriction of \(T\) to each region \(X^{n}\times R_{\sigma}\) is independent of \(v\in R_{\sigma}\), and it reduces to a linear operator:
\begin{equation}
  T((x_{i}), (v_{i})) = (x_{\sigma(1)}, \dots, x_{\sigma(k)}).
\end{equation}

It follows that \(T\) is differentiable in the interior of each region \(X^{n}\times R_{\sigma}\), and its non-differentiability points are constrained to the boundaries of the differentiability regions, i.e. the set \(X^{n} \times D\), where \(D = \{x\in\RR^{n}: x_{i}=x_{j} \, \text{for some $i\ne j$}\}\).

In particular, since D is a union of hyperplanes of codimension \(1\) in \(\RR^{n}\), the non-differentiability set of \(T\) has measure \(0\). Just as in the simpler case of the \(\mathrm{ReLU}\) activation function, the non-differentiability of the hard top-\(k\) operator is not a serious problem---which is a possible misconception here.

The real problem is that although the gradient of \(T\) exists (almost everywhere), it is not particularly useful, since
\begin{equation}
  \frac{\partial T}{\partial v_{i}} = 0,
\end{equation}
because in each region \(X^{n}\times R_{\sigma}\) the operator \(T\) is independent of \(v_{i}\). This makes back-propagation through the scores impossible, and prevents training the scoring function. It could be seen as an extreme case of the vanishing gradient problem.
In the next section, we introduce a mechanism not prone to this issue.

\begin{algorithm}
    \caption{Successive Halving Top-$k$ Selection}
    \begin{algorithmic}[1]
    \Procedure{TopK}{$E, v$}
    \For{$i \gets 1, \log_2(\lceil n/k \rceil)$}
    \State $E, v \gets \Call{Sort}{E, v}$
    \State $E, v \gets \Call{Tournament}{E, v}$
    \EndFor
    \State \textbf{return} $E$
    \EndProcedure\\

    \Procedure{Sort}{$E, v$}
    \State $v' \gets (v_1,v_2,..), $ where $v_i \ge v_{i+1}$ and $v_i \in v$
    \State $E' \gets (E_1,E_2,..), $ where $v_i \ge v_{i+1}$ and $v_i \in v$ 
    \State \textbf{return} $E', v'$
    \EndProcedure\\
    
    \Procedure{Tournament}{$E, v$}
    \State $n  \gets \frac{1}{2} \lVert v \rVert$ \Comment{Target size}
    \State $d  \gets \lVert E_{*,1} \rVert$ \Comment{Representation depth}
    \State $v' \gets 0_{n,1}$  
    \State $E' \gets 0_{n,d}$
    \For{$i \gets 1, n$}
        \State $w \gets \Call{PeakedSoftmax}{v_i, v_{2n-i+1}}$
        \State $E'_i \gets E_i \cdot w_0 + E_{2n-i+1} \cdot w_1$
        \State $v'_i \gets v_i \cdot w_0 + v_{2n-i+1} \cdot w_1$
    \EndFor
    \State \textbf{return} $E', v'$
    \EndProcedure
    \end{algorithmic}
    \label{algo-soft-topk}
\end{algorithm}

\subsection{Analysis and Discussion}
We propose an $\mathcal{O}(n\log_2(n/k))$ time-complexity algorithm for selecting $k$ top-scoring representations from a vector of length $n$. An iterative approach of \citet{goyal2017continuous} with $\mathcal{O}(nk)$ complexity involves a higher cost for almost any $k$. The total number of exponentiation operations in the Successive Halving Top-$k$ is bounded by $2n$, as each round of the tournament halves the input size. Compared to $kn$ in the case of the \citet{goyal2017continuous} algorithm, orders of magnitude savings in expensive exponentiation operations are obtained. 

Another key requirement for a robust top-$k$ algorithm is to accurately approximate hard selection.
Meanwhile, iteration-based algorithm disperses the probability mass over all items, resulting in a poor approximation of top-$k$. This inefficiency of softmax over long vectors can be overcome by multiplying them by a large constant; however, this leads to numerical instability. 
Moreover, they tend to perform worse when employed as a neural network layer due to the long chain of backpropagation's dependencies.

In contrast, we always perform softmax over a pair of values, guaranteeing that there will be a candidate with a $\ge0.5$ probability assigned. After each pass, the best scoring $k$ vectors with a small noise are obtained. It is a result of interpolating with the lower-scoring element from each pair. 

As stated in the paper, we ensure that strong candidates have weakly-scoring opponents, strengthening their presence in the tournament's next round. The fundamental requirement of this trick is to sort inputs, resulting in an additional cost of $\mathcal{O}(n \log(n))$. However, in the case of modern CPUs, this cost is practically negligible. Yet, the sorting step can be omitted, leading to a slightly degraded top-$k$ approximation. 
During the process, a vector with considerable noise may be produced for elements with indexes closer to the $n/2$. Nevertheless, some noise itself is desired, as it allows gradients to propagate to elements out of the top-$k$.

\subsection{Differential Properties\label{sec:k2properties}}

Recall the description of hard top-$k$ from Section~\ref{sec:limitations}. The main advantage introduced by soft top-\(k\) operator of Successive Halving, is providing reasonable gradients with respect to the scores \(v_{i}\). This allows to create a \emph{trainable} pooling mechanism reducing the number of output embeddings. At the same time, it does not improve differentiability---which is another possible misconception we wanted to dispel.


In our proposed approach we assume that both \(n\) and \(k\) are powers of \(2\). The soft top-\(k\) operator is then defined through a composition of \(\log_{2}(n/k)\) \emph{halving operators} \(H_{n}\colon X^{n}\times\RR^{n} \to X^{n/2}\times \RR^{n/2}\), reducing the number of vectors and their scores by half (see Appendix~\ref{appendix_topk}).

The halving operator itself is the composition of sorting the vectors together with their scores, and a transformation \(C\colon X^{n}\times\RR^{n}\to X^{n/2}\times\RR^{n/2}\) producing \(n/2\) convex combinations of the form 
\begin{equation}
    \label{eq:convex-op-1}
  y_{i} = w_{i} x_{i} + (1-w_{i}) x_{n+1-i},
\end{equation}
where the weights are the softmax of the pair of scores \((v_{i}, v_{n+1-i})\), i.e.
\begin{equation}
    \label{eq:convex-op-2}
    w_{i} = \frac{e^{v_{i}}}{e^{v_{i}} + e^{v_{n+1-i}}}.
\end{equation}

Similarly as in the case of the hard top-\(k\) operator, the non-differentiability of \(H_{n}\) arises from sorting. The convex combinations however smooth out some of the non-differentiabilities.

Let \(\tau\in S_{n}\) be the transposition of \(i\) and \(n+1-i\). The transformation \(C\) is then invariant under \(\tau\), which transposes both the weights \((w_{i}, 1-w_{i})\), and vectors \((x_{i}, x_{n+1-i})\). Hence, \(C\) is invariant under the subgroup \(G\subseteq S_{n}\) generated by such transpositions. As a consequence, on the set \(X^{n}\times \bigcup_{\rho\in G\sigma}R_{\rho}\) the operator \(H\) is given by
\begin{multline}
  H_{n}((x_{i}), (v_{i})) =\\= C((x_{\sigma(1)}, \dots, x_{\sigma(n)}), (v_{\sigma(1)}, \dots, v_{\sigma(n)})),
\end{multline}
and since \(C\) is differentiable, so is the restriction of \(H\) to this region.

In summary, while in the case of the hard top-\(k\) operator there are \(n!\) differentiability regions corresponding to sorting permutations, for the halving operator the differentiability regions are their unions corresponding to the cosets of \(G\) in \(S_{n}\). Since the generating transpositions of \(G\) are disjointly supported, it is isomorphic to \(\ZZ_{2}^{n/2}\), and therefore there are \(2^{-n/2}n!\) differentiability regions.

The Successive Halving top-$k$ operator is the composition of multiple halving operators, each introducing new non-differentiabilities, and the final projection onto $X^{k}$. The arising non-differentiability set is still of measure $0$, which is covered in detail in Appendix~\ref{appendix_math}.

\subsection{Differential Properties of Complete Successive Halving Top-\textit{k} Operator}\label{appendix_math}

We have shown that hard top-$k$ operator makes back-propagation through the scores impossible, and prevents training the scoring function (Section~\ref{sec:limitations}), whereas top-$\tfrac{n}{2}$ halving is not prone to this problem (Section~\ref{sec:k2properties}). We discuss the properties of full-featured \emph{Successive} Halving bellow.

We have previously covered the case of \(H_{n}\). But the succesive halving top-\(k\) operator \(\Gamma\colon X^{n}\times\RR^{n} \to X^{k}\) is the composition 
\begin{equation}
\label{eq:composition}
    \Gamma = \operatorname{pr}_{X^{k}} \circ H_{2k}\circ H_{4k} \circ\dots\circ H_{n/2} \circ H_{n}
\end{equation}
of multiple halving operators, each introducing new non-differentiabilities, and the projection \(\operatorname{pr}_{X^{k}}\colon X^{k}\times\RR^{k}\to X^{k}\). The non-differentiability set of \(\Gamma\) is contained in the preimages of non-differentiability sets of the \(H_{i}\) with respect to the preceding factors in the composition.

In such a situation it is generally not obvious that the resulting non-differentiability set is still of measure \(0\). To remedy this, let us first make some general observations about differentiability sets of mappings between manifolds. 

For a mapping $F\colon M\to N$ of smooth manifolds, denote by $Z_F$ the set of all points $p\in M$ such that either $F$ is not smooth in any neighborhood of $p$, or the rank of the derivative of $F$ at $p$ is not maximal. Observe that if the closure $\overline{Z_F}$ of $Z_F \subseteq M$ has measure $0$, then the preimage $F^{-1}[E]$ of any set $E\subset N$ of measure $0$ is itself of measure $0$. Indeed, we may decompose such preimage as
\begin{multline}
    F^{-1}[E] = (F^{-1}[E] \cap \overline{Z_F}) \cup\\\cup (F^{-1}[E] \cap (M\setminus\overline{Z_F})),
\end{multline}
where the first component has measure zero (being a subset of $\overline{Z_F}$), while the second component can be covered by a countable family of open sets on which $F$ is differentiable, its derivative has maximal rank, and the constant rank theorem applies. Thus, locally on each set $U$ of this cover, $F$ is conjugate to a projection $\RR^m \to \RR^n$, and $F|_U^{-1}[E]$ has measure 0. In the end, $F^{-1}[E]$ is decomposed into a countable union of zero-measure sets, so it has measure $0$.

It follows that if $G\colon N\to P$ is another mapping such that $\overline{Z_G}$ has measure $0$ in $N$, then $\overline{Z_{G\circ F}}$ also has measure 0, since
\begin{multline}
    \overline{Z_{G\circ F}} \subseteq \overline{Z_F \cup F|_{M\setminus Z_F}^{-1}[Z_G]} =\\=\overline{Z_F} \cup F|_{M\setminus Z_F}^{-1}[\overline{Z_G}].
\end{multline}

Above, $F|_{M\setminus Z_F}^{-1}$ commutes with the closure operator because the restriction $F|_{M\setminus Z_F}$ is continuous. This result extends by induction to compositions of any number of mappings.

In order to show that $\Gamma$ defined as the composition \eqref{eq:composition} is almost everywhere differentiable, it therefore suffices to prove that $Z_{\Gamma}$ has measure 0, which in turn amounts to showing that $\overline{Z_{H_i}}$ has measure zero for any halving transformation $H_i$. Recall that the halving transformation is the composition of the corresponding sorting operator and convex combination operator $C$ defined in \eqref{eq:convex-op-1} and \eqref{eq:convex-op-2}. 

For the sorting operator, the non-differentiability set is a union of a finite number of hyperplanes, hence a closed set of measure zero, and outside this set the derivative has maximal rank. The operator $C$ on the other hand is smooth, and it remains to verify the rank of its derivative. Denote \(((y_{i}), (u_{i})) = C((x_{i}), (v_{i}))\), and observe that \(\partial u_{i}/\partial x_{j} = 0\). Therefore it is enough to show that the matrices of partial derivatives \((\partial y_{i}/\partial x_{j})_{ij}\) and \((\partial u_{i}/\partial v_{j})_{ij}\) have linearly independent columns.
For \(j\in\{i, 2m+1-i\}\) we have
\begin{equation}
  \frac{\partial y_{i}}{\partial x_{j}} = \frac{e^{v_{j}}}{e^{v_{i}} + e^{v_{2m+1-i}}} > 0,
\end{equation}
and \({\partial y_{i}}/{\partial x_{j}}=0\) for all other \(j\). Since the sets \(\{i, 2m+1-i\}\) are pairwise disjoint, the columns are linearly independent.

In case of \(\partial u_{i}/\partial v_{j}\) the reasoning is similar. They are again nonzero only for  \(j\in\{i, 2m+1-i\}\), for which
\begin{multline}
  \frac{\partial u_{i}}{\partial v_{j}} =\\=\frac{e^{v_{j}}\bigl( e^{v_{2m+1-j}}(v_{j}-v_{2m+1-j}) + e^{v_{j}} + e^{v_{2m+1-j}} \bigr)}{(e^{v_{j}}+e^{v_{2m+1-j}})^{2}} ,\\
\end{multline}
and this is strictly positive for at least one \(j\in\{i, 2m+1-i\}\). It follows that the columns are non-zero and have non-zero entries in different rows, so again they are linearly independent.

We have therefore shown that the Jacobian matrix of \(C\) has linearly independent columns, or in other words, its derivative is surjective at every point, which is what we needed to complete the proof that the non-differentiability set of \(\Gamma\) can be covered by a locally finite family of codimension 1 submanifolds, thus being of measure \(0\).

\subsection{Performance}\label{topk_performance}
\begin{figure}
    \centering
    \includegraphics[width=\linewidth,trim={0.3cm 0.5cm 0.4cm 0.4cm},clip]{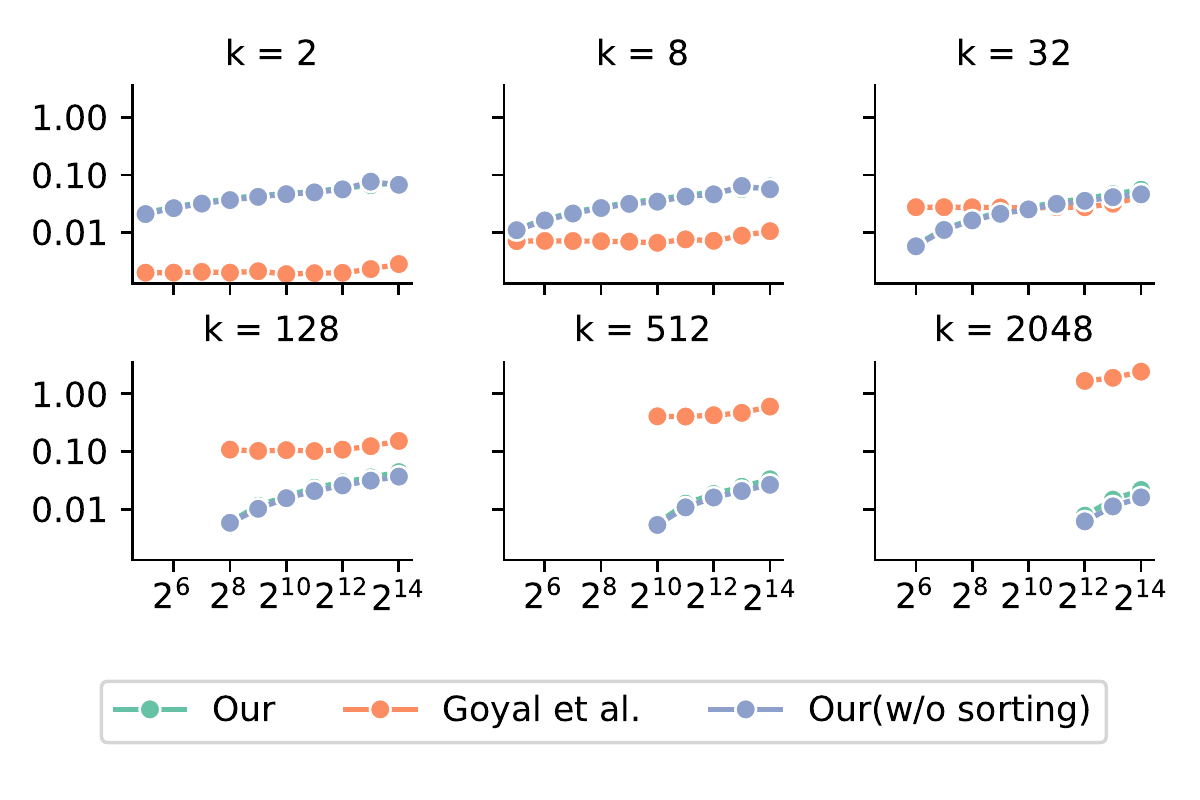}
    \caption{Number of seconds required to process a batch of sequences ($Y$-axis). The lower the better. Results depending on $n$ ($X$-axis) for various values of $k$, assuming $k<n$. Depicted solution without sorting partially covers the data points of the solution with sorting(Our).}
    \label{fig:performance}
\end{figure}

\begin{figure}
    \centering
    \includegraphics[width=\linewidth,trim={0.7cm 0.5cm 0.3cm 0.4cm},clip]{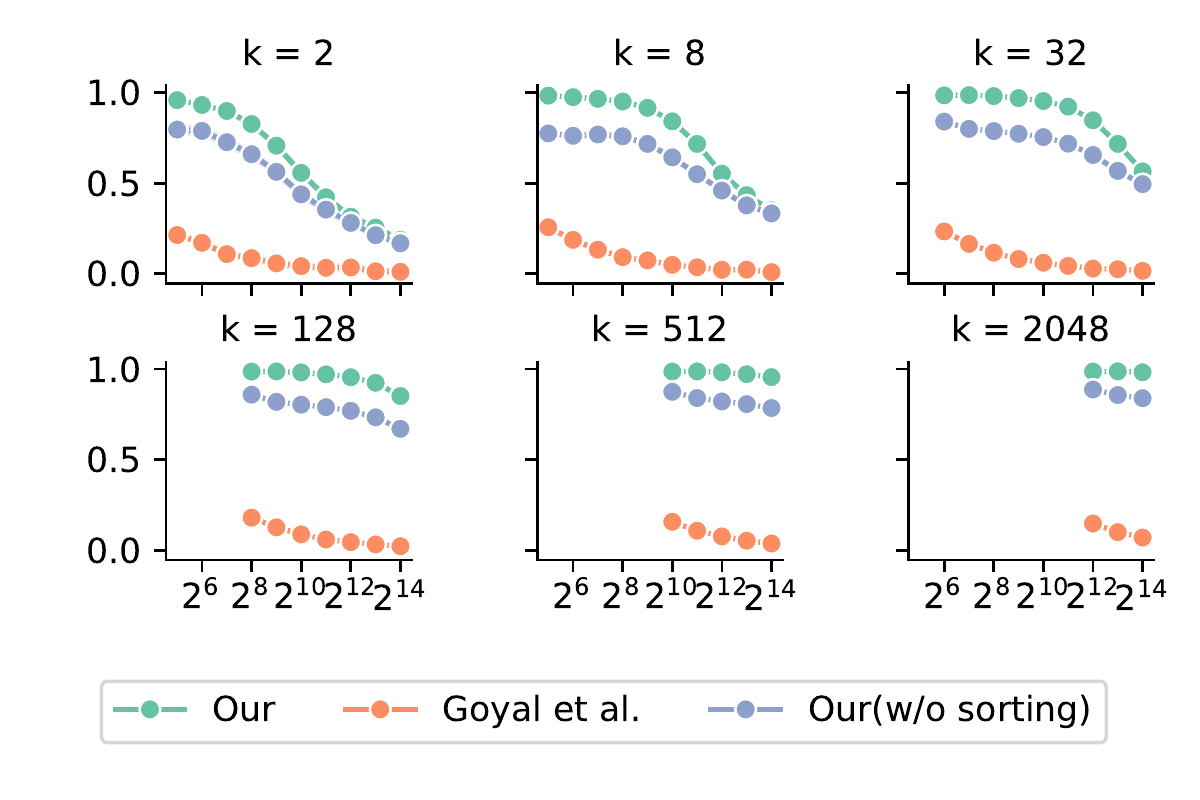}
    \caption{Approximation quality ($Y$-axis) in the $nCCS$ metric. The higher the better. Results depending on $n$ ($X$-axis) for various values of $k$, assuming $k<n$.}
    \label{fig:performance2}
\end{figure}

In Figure~\ref{fig:performance} and \ref{fig:performance2} we show that our approach is highly similar to real top-$k$ for any given $k$, and is significantly faster than alternative solutions, such as, e.g., iterative top-$k$ selection.

We assessed the performance of the Successive Halving Top-$k$ as compared to \citet{goyal2017continuous} experimentally, on randomly sampled matrices $E$ such that $\mathnormal{E_{ij}\sim \mathcal{U}}[-1,1]$ and scores $\mathnormal{v_{i}\sim \mathcal{U}}[0,1]$. The selected $k$ top-scoring vectors were compared to the real top-$k$ selection using normalized Chamfer Cosine Similarity (nCCS) as given:
\begin{equation*}
nCCS = \frac{1}{k} \sum_{i =1}^{k} \max_{j \in [1,k]}(\operatorname{cos}( y_i, \hat{y_j}))
\end{equation*}

Additionally, we measured an average time for processing a batch of size $16$ on the NVIDIA A100 GPU,
and addressed the question of how both algorithms differ in terms of speed (Figure \ref{fig:performance}) and quality (Figure \ref{fig:performance2}), depending on $k$ and $n$ choices.
One can notice that the higher the choice of $k$, the faster our algorithm is, and the slower is the iterative baseline of \citet{goyal2017continuous} as predicted by their complexities. Our solution's qualitative robustness is proven by achieving higher similarity to real top-$k$ for any given $k$. The score degrades as the number of rounds in the tournament increases, as each round introduces additional noise.\\
To assess the importance of the sorting step, we removed it from the algorithm and compared with the proposed top-k. The results suggests that sorting is efficient and fast, as it is introduces average time overhead of 7.3\%, while allowing error to be reduced by 45.2\% on average.

\section{Summarization Experiments}\label{appendix_experiment_details}

This appendix covers other ablation studies and details of previously-reported experiments.

\subsection{Shallow Models Setup}

{
\paragraph{Shared setup.}
The models were trained using the Adam optimizer and cross-entropy loss, with hyperparameters specified in Table~\ref{tab:hp}. Validation was performed every three epochs on a validation set and the training stopped when no progress was observed taking the seven last scores into account. Presented scores are the best scores on a validation set. All of the considerations assumed the use of dot-product attention except for LSH and Efficient Transformers.
\begin{table}
\caption{Hyperparameters for shallow models used in the summarization experiments.}
\label{tab:hp}
\centering
\begin{tabular}{lc}
    \toprule
    Hparam & Value \\
    \midrule
    Encoder Layers & 2 \\
    Decoder Layers & 2 \\
    Vocab size & 32k \\
    Dropouts  & .1    \\
    Activation & ReLU \\
    Emb dim & 512 \\
    FFN emb dim & 2048  \\
    Encoder positional emb & sinusoidal \\
    Decoder positional emb & None \\
    Batch size          & 256 \\
    Learning rate       & 5e-4 \\
    Learning rate decay & -- \\
    Shared emb & True \\
    Weight decay        & .1 \\
    Attention heads & 8 \\
    Beam size & 8 \\
    Total parameters & ~32M \\
    \bottomrule
\end{tabular}
\end{table}

\begin{table}
\caption{Hyperparameters for DeepPyramidion and deep Blockwise baseline models used in the summarization experiments.}
\label{tab:hp_deep}
\centering
\begin{tabular}{lc}
    \toprule
    Hparam & Value \\
    \midrule
    Encoder Layers & 6 \\
    Decoder Layers & 6 \\
    Vocab size & 32k \\
    Dropouts  & .1    \\
    Activation & ReLU \\
    Emb dim & 768 \\
    FFN emb dim & 3072  \\
    Encoder positional emb & sinusoidal \\
    Decoder positional emb & None \\
    Batch size          & 256 \\
    Learning rate       & 5e-4 \\
    Learning rate decay & -- \\
    Shared emb & True \\
    Weight decay        & .1 \\
    Attention heads & 8 \\
    Warmup steps & 5k \\
    Total Parameters & ~124M \\
    \bottomrule
\end{tabular}
\end{table}

\paragraph{Vanilla.}
The exact setup of Vanilla Transformer is provided in Table~\ref{tab:hp}.

\paragraph{Blockwise.}
We employed block attention with window size and stride equal to 512. We use block attention in the encoder, and the decoder features dense attention.
The rest of the parameters follows shared setup.

\paragraph{Transpooler.}
Transpooler features linear scorer and successive halving algorithm. It uses Blockwise's setup of blockwise attention. Pooling is performed after the last encoder layer.
The number of halving rounds depends on the proportion of maximal input sequence size and the desired bottleneck size.
Transpoolers models were trained and validated with our soft top-$k$.

In the case of input chunking and use of blockwise attention, positions were calculated originating at the beginning of document. For simplicity, no positional embeddings were used on the decoder side. We argue, that embeddings passed down have already sufficient positional information from the encoder. 
\paragraph{LSH.}
All of the previous considerations assumed the use of dot-product attention with memory and computational costs growing quadratically with the input size.
Baselines relying on either efficient or LSH-based attention were conducted with two heads of local window attention that has been shown to improve models with long-range sparsity \citep{rae-razavi-2020-transformers}. Without local attention, their results were several points lower. We assumed an LSH bucket size of 64 and four parallel hashes. Bucket size follows the authors' recommendations, whereas the number of hashes is a reasonable trade-off between memory complexity and approximation quality \citep{Kitaev2020ReformerTE}. Although one may obtain slightly better scores with eight hashes, it would result in higher memory consumption than in the case of full attention baselines for all of the considered sequence lengths. The rest of the parameters follow the Blockwise baseline.

\paragraph{Efficient Transformer.}
The training setup follows the original work. The Efficient Transformer does not have any specific parameters to determine, so all other training/validation choices agree with Blockwise baseline. 

\paragraph{Funnel Transformer.}
The training setup of Funnel follows the original work, with the specific strided mean pooling and upsampling before passing to the decoder. For example, in Funnel $8k \rightarrow 512$ (pooling from 8k to 512), 16 consecutive tokens were averaged after the first encoder layer. The decoder size is 8k, and the residual connections start from the first's layer output (taken just before pooling).

\paragraph{PoWER-BERT.}
As it comes to the PoWER-based models, we finetune Vanilla transformers with a progressive elimination of word vectors on the encoder side, following the approach of \citet{pmlr-v119-goyal20a}. We do not optimize the number of eliminated embeddings but assume the fixed reduction, similarly to our Pyramidion models.
Additionally, Table~\ref{tab:length} reports results with a progressive elimination of word vectors on the encoder side, adapted from PoWER-BERT \citep{pmlr-v119-goyal20a}. Note that models are not trained from scratch in this approach, and we assumed blockwise attention to make it comparable with our models (see Appendix~B). We started from appropriate checkpoints of a blockwise model and finetuned it for ten epochs. Here, we validated every one epoch. As training time, we provide times achieved during this finetuning. As presumed, a hard selection of word vectors offers an improved inference time for the cost of slightly decreased ROUGE scores. }

\subsection{Number of Layers, Bottleneck Size}\label{appendix_sub_pyramidion}
Deeper Pyramidion and Transpooler models with various pooling configurations were further examined in Table~\ref{tab:pyramidion}. The training setup follows the previously described Transpooler setup. In the case of Pyramidion, we pool after the first or the second layer in the encoder. 
Scores of Pyramidion with pooling operation after the second and subsequent layers are significantly higher than \#9, presumably because the representations after the first layer are not reliable enough to produce meaningful scores.

The Pyramidion with a three-layer encoder that reduces the input of $8k$ tokens gradually to $2k$ [\#13] offers results $1.2$ points better than the Vanilla model consuming input of the same length [\#3]. Additionally, the complexity was reduced by a factor of $13$ and $4$ in the encoder and decoder, respectively, while achieving $3\times$ training and $2.4\times$ inference acceleration.

Finally, a series of Pyramidion experiments confirmed the applicability of gradual pooling with bottlenecks of $128$, $512$, and $2$k sizes [\#12, \#11, \#13]. It can be noticed that a reduction in the bottleneck's size leads to a decrease in performance.

\begin{table*}
    \caption{Scores and complexities of the Pyramidion and Transpooler with different encoder and decoder depths, as well as various lengths after pooling. The input of 8k representations pooled gradually to decoder length. Two-layer decoder and encoder of depth ranging from $2$ to $4$ layers. Arrow $\rightarrow$ denotes an additional pooling between encoder layers.}
    \label{tab:pyramidion}
    \centering
    \begin{tabular}{lllrrrrrr}
    \toprule
    \multirow{2}{*}{\#} &
    \multirow{2}{*}{Architecture} &
    \multicolumn{2}{c}{Lengths} &
    \multicolumn{2}{c}{Time} &
    \multicolumn{2}{c}{ROUGE} \\
    & & Encoder & Decoder & Training & Inference & R-1 & R-2 \\
    \midrule
    21 & \ldelim\{{4}{50pt}[Pyramidion ] & 8k $\rightarrow$ 2k & 512 & 1.07 & 4.18 & 31.1 & 11.5 \\
    22 & & 8k, \quad 8k $\rightarrow$ 2k & 512 & 1.55 & 4.26 & 41.2 & 16.5 \\
    23 & & 8k, \quad 8k $\rightarrow$ 2k $\rightarrow$ 512 \hspace{-1cm} & 128 & 1.78 & 3.74 & 37.3 & 14.3 \\ \vspace{0.1cm}
    24 & & 8k, \quad 8k $\rightarrow$ 4k & 2k & 1.47 & 5.49 & \textbf{43.0}& \textbf{17.2} \\
    25 & \ldelim\{{1.8}{40pt}[Transpooler ] & 8k, \quad 8k & 2k & 1.26 & 5.51 & 42.7 & 16.7\\
    26 & & 8k, \quad 8k, \quad 8k & 2k & 1.74 & 5.54 & \textbf{43.1} & \textbf{17.3} \\  
    \bottomrule
    \end{tabular}
\end{table*}

\begin{table*}
\caption{Scores depending on blockwise attention block size and sparsification mechanism with $2k$ and $8k$ encoder input length considered. Different models with a two-layer encoder and a two-layer decoder.}
\label{tab:blocksize}
\centering
\setlength{\tabcolsep}{5pt}
\begin{tabular}{rcrrrrrrr}
    \toprule
    \multirow{2}{*}{\#} &
    \multirow{2}{*}{Pooling} &
    \multirow{2}{*}{Block size} &
    \multicolumn{2}{c}{Lengths} &
    \multicolumn{2}{c}{Time} &
    \multicolumn{2}{c}{ROUGE} \\
    & & & Encoder & Decoder & Training & Inference & R-1 & R-2 \\
    \midrule
    
    27 & \ldelim\{{3}{50pt}[No pooling ] & 128 & 2k & 2k & 0.25 & 5.11 & \textbf{39.1} & \textbf{14.4} \\
    28 & & 512 & 2k & 2k & 0.31 & 5.28 & 38.6 & 14.1 \\ \vspace{0.1cm}
    29 & & \textit{(without)} & 2k & 2k & 0.60 & 5.77 & 38.2 & 14.0 \\
    30 & \ldelim\{{3}{55pt}[Transpooler ] & 128 & 2k & 512 & 0.49 & 3.99 & 38.2 & 14.1 \\
    31 & & 512 & 2k & 512 & 0.54 & 4.24 & \textbf{39.1} & \textbf{14.6} \\ \vspace{0.1cm}
    32 & & \textit{(without)} & 2k & 512& 0.82 & 4.49 & 37.1 & 13.7 \\
    \bottomrule
    \end{tabular}
\end{table*}

\subsection{Effect of Block Size}
We provide ablation experiments on block size effects in Table~\ref{tab:blocksize}. For simplicity, all of the previous experiments were conducted with an attention block size of $512$ where applicable.
Block consisting of $128$ tokens lead to an improved encoder complexity and slightly lower computation time [\#25, \#28, \#31]. It is not always achieved at the price of decreased ROUGE scores.

The scoring mechanism introduces some overhead during the training, which may be noticeable for shorter sequences.
However, when it comes to the inference time we aimed at when proposing the method, it can be observed that a pooling operation positively impacts it. Pooling improves the inference time whether or not it is used in combination with blockwise attention.

\section{Effect of Input Length}
The importance of the longer input for the overall performance can be deduced by analyzing the performance of models \#1-\#8 in Table~\ref{tab:length}, where we employed different input lengths for different models (Vanilla, Blockwise, and Pyramidion), and found out that a steady gain of $3.3-3.6$ R1 (and $2.1-2.6$ R2) points is observed for all of them when the input length is extended from $2k$ to $8k$. Please note that while these results are provided in the ablation study that features a shallower network, the difference is significant and consistent. Hence, we did not repeat the experiment in the deeper setup.

\subsection{Deep Model Setup}\label{appendix_sub_deeper}
\textbf{Training.}
Table~\ref{tab:hp_deep} presents the shared setup of a DeepPyramidion and Blockwise, evaluated in the Section~\ref{sec:evaluation}. We train until the validation score was not achieved for $7$ consecutive validations.

\paragraph{Inference.}
We follow parameters for the generation of HAT-BART \cite{rohde2021hierarchical}: a beam width of $2$, length penalty of $1$, and minimum and maximum generation lengths of $72$ and $966$, respectively. We validated on the validation set every three epochs and chose the best performing model to generate outputs on the test set. 

\subsection{Hardware and Software Used}\label{appendix_sub_hyperparams}
All experiments and benchmarks were performed on a DGX-A100 server equipped with eight NVIDIA Tesla A100 GPUs. We based our experiments using \textit{fairseq} \citep{ott-etal-2019-fairseq} v0.9.0, Python 3.6.10, PyTorch 1.6.0a0+9907a3e \citep{NEURIPS2019_9015}, CUDA Version 11.0 and NVIDIA drivers 450.51.06. We trained in a full precision.

\subsection{Detailed Results}

\begin{table}[b]
    \caption{Scores with 95\% bootstrap confidence intervals of an estimate of the data \citep{doi:10.1113/jphysiol.2012.239376}.}
    \label{tab:bootstrap}
    \centering
    \begin{tabular}{rrrrr}
    \toprule
    \# & \multicolumn{2}{l}{ROUGE-1 (CI)} & \multicolumn{2}{l}{ROUGE-2 (CI)} \\
    \midrule
    1 & $28.1$ & $(27.8-28.3)$ & $8.3$ & $(8.1-8.4)$ \\
    2 & $38.2$ & $(37.9-38.5)$ & $14.0$ & $(13.8-14.2)$ \\
    3 & $41.8$ & $(41.6-42.1)$ & $16.1$ & $(15.9-16.4)$ \\
    4 & $38.6$ & $(38.3-38.8)$ & $14.1$ & $(13.9-14.3)$ \\
    5 & $41.9$ & $(41.6-42.1)$ & $16.7$ & $(16.5-17.0)$ \\
    6 & $39.1$ & $(38.9-39.4)$ & $14.6$ & $(14.4-14.8)$ \\
    7 & $41.8$ & $(41.6-42.1)$ & $16.4$ & $(16.2-16.7)$ \\
    8 & $42.7$ & $(42.4-43.0)$ & $16.7$ & $(16.5-16.9)$ \\
    9 & $28.5$ & $(28.3-28.7)$ & $7.5$ & $(7.4-7.6)$ \\
    10 & $33.6$ & $(33.4-33.8)$ & $10.5$ & $(10.4-10.6)$ \\
    11 & $35.7$ & $(35.5-36.0)$ & $11.2$ & $(11.1-11.4)$ \\
    12 & $28.4$ & $(28.2-28.6)$ & $7.8$ & $(7.7-7.9)$ \\
    13 & $34.1$ & $(33.9-34.4)$ & $10.4$ & $(10.3-10.6)$ \\
    14 & $35.0$ & $(34.7-35.2)$ & $10.8$ & $(10.7-11.0)$ \\
    15 & $35.3$ & $(35.0-35.5)$ & $12.7$ & $(12.5-12.9)$ \\
    16 & $36.9$ & $(36.6-37.2)$ & $14.1$ & $(13.9-14.4)$ \\
    17 & $42.0$ & $(41.7-42.3)$ & $16.5$ & $(16.3-16.7)$ \\
    18 & $38.6$ & $(38.3-38.8)$ & $14.3$ & $(14.1-14.5)$\\
    19 & $41.8$ & $(41.6-42.1)$ & $16.5$ & $(16.3-16.8)$\\
    20 & $42.0$ & $(41.7-42.2)$ & $16.4$ & $(16.2-16.6)$\\
    
    21 & $31.1$ & $(30.7-31.6)$ & $11.5$ & $(11.3-11.7)$ \\
    22 & $41.2$ & $(40.9-41.4)$ & $16.5$ & $(16.3-16.8)$ \\
    23 & $37.3$ & $(37.1-37.6)$ & $14.3$ & $(14.1-14.5)$ \\
    24 & $43.0$ & $(42.7-43.3)$ & $17.2$ & $(17.0-17.5)$ \\
    25 & \multicolumn{4}{l}{\textit{$\rightarrow$ See  \#8}} \\
    26 & $43.1$ & $(42.8-43.3)$ & $17.3$ & $(17.0-17.5)$ \\

    27 & $39.1$ & $(38.8-39.3)$ & $14.4$ & $(14.2-14.6)$ \\
    28 & $38.6$ & $(38.3-38.8)$ & $14.1$ & $(13.9-14.3)$ \\
    29 & \multicolumn{4}{l}{\textit{$\rightarrow$ See  \#2}} \\
    30 & $38.2$ & $(38.0-38.4)$ & $14.1$ & $(13.9-14.3)$ \\
    31 & \multicolumn{4}{l}{\textit{$\rightarrow$ See  \#6}} \\
    32 & $37.1$ & $(36.9-37.4)$ & $13.7$ & $(13.5-13.8)$ \\

    \bottomrule
    \end{tabular}
\end{table}

\begin{table}
    \caption{Mean time of processing and inference in seconds $\pm$ standard deviation. We assumed a fixed length of $256$ or $512$ tokens to decode to discount for lower processing time of models predicting the end of sequence token earlier.}
    \label{tab:timeci}
    \small
    \centering
    \begin{tabular}{rrrr}
    \toprule
    \# & Training & Inference @ 256 & Inference @ 512 \\
    \midrule
    1 & $0.13$ $\pm 0.02$ & $2.05$ $\pm 0.01$ & $4.23$ $\pm 0.01$ \\ 
    2 & $0.60$ $\pm 0.03$ & $2.76$ $\pm 0.01$ & $5.77$ $\pm 0.02$ \\ 
    3 & $4.46$ $\pm 0.26$ & $6.56$ $\pm 0.03$ & $13.27 \pm 0.06$ \\ 
    4 & $0.31$ $\pm 0.02$ & $2.58$ $\pm 0.00$ & $5.28 \pm 0.01$ \\ 
    5 & $0.85$ $\pm 0.12$ & $5.40$ $\pm 0.00$ & $11.49 \pm 0.01$ \\ 
    6 & $0.54$ $\pm 0.02$ & $2.09$ $\pm 0.00$ & $4.24 \pm 0.01$ \\ 
    7 & $1.44$ $\pm 0.04$ & $2.14$ $\pm 0.00$ & $4.28 \pm 0.01$ \\ 
    8 & $1.26$ $\pm 0.06$ & $2.71$ $\pm 0.00$ & $5.51 \pm 0.01$ \\ 
    9 & $0.19$ $\pm 0.02$ & $2.16$ $\pm 0.01$ & $4.27 \pm 0.01$ \\
    10 & $0.56$ $\pm 0.03$ & $3.01$ $\pm 0.01$ & $5.92 \pm 0.01$ \\
    11 & $1.69$ $\pm 0.12$ & $0.87$ $\pm 0.05$ & $13.41 \pm 0.07$ \\
    12 & $0.12$ $\pm 0.02$ & $2.16$ $\pm 0.01$ & $4.20 \pm 0.01$ \\
    13 & $0.29$ $\pm 0.03$ & $2.98$ $\pm 0.02$ & $5.91 \pm 0.01$ \\
    14 & $0.82$ $\pm 0.10$ & $6.91$ $\pm 0.06$ & $13.75 \pm 0.08$ \\
    15 & $1.04$ $\pm 0.04$ & $2.17$ $\pm 0.11$ & $4.28 \pm 0.18$ \\
    16 & $1.87$ $\pm 0.16$ & $2.71$ $\pm 0.09$ & $5.33 \pm 0.15$ \\
    17 & $2.06$ $\pm 0.16$ & $3.57$ $\pm 0.12$ & $6.92 \pm 0.17$ \\
    18 & $0.61$ $\pm 0.11$ & $2.07$ $\pm 0.06$ & $4.01 \pm 0.04$ \\
    19 & $1.78$ $\pm 0.14$ & $2.08$ $\pm 0.07$ & $4.03 \pm 0.06$ \\
    20 & $1.53$ $\pm 0.13$ & $2.64$ $\pm 0.07$ & $5.25 \pm 0.04$ \\
    21 & $1.05$ $\pm 0.05$ & $2.12$ $\pm 0.01$ & $4.18 \pm 0.01$ \\
    22 & $1.55$ $\pm 0.04$ & $2.12$ $\pm 0.01$ & $4.26$ $\pm 0.01$  \\ 
    23 & $1.78$ $\pm 0.05$ & $1.86$ $\pm 0.01$ & $3.74$ $\pm 0.01$  \\ 
    24 & $1.47$ $\pm 0.04$ & $2.69$ $\pm 0.01$ & $5.49$ $\pm 0.01$  \\ 
    25 & \multicolumn{3}{l}{\textit{$\rightarrow$ See \#8}} \\
    26 & $1.74$ $\pm 0.05$ & $2.73$ $\pm 0.01$ & $5.54$ $\pm 0.01$  \\ 

    27 & $0.25$ $\pm 0.02$ & $2.51$ $\pm 0.00$ & $5.11 \pm 0.01$ \\ 
    28 & $0.31$ $\pm 0.02$ & $2.58$ $\pm 0.00$ & $5.28 \pm 0.01$ \\ 
    29 & \multicolumn{3}{l}{\textit{$\rightarrow$ See \#2}} \\
    30 & $0.49$ $\pm 0.03$ & $2.04$ $\pm 0.01$ & $3.99 \pm 0.01$ \\ 
    31 & \multicolumn{3}{l}{\textit{$\rightarrow$ See  \#6}} \\
    32 & $0.82$ $\pm 0.03$ & $2.20$ $\pm 0.01$ & $4.49 \pm 0.02$ \\ 

    \bottomrule
    \end{tabular}
\end{table}

Table~\ref{tab:bootstrap} reports ROUGE scores for all of the evaluated models. In addition, we report $95\%$ bootstrap confidence intervals of an estimate of the data here to mean scores.

The average time of processing a batch of documents is reported in Table~\ref{tab:timeci}. We used batch of size $64$ for training, and $8$ for inference. Decoding experiments were synthetic. Specifically, we assumed a fixed length of either $256$ or $512$ tokens to decode to discount for lower processing time of models predicting the end of sequence token earlier.

\end{document}